# VMAV-C: A Deep Attention-based Reinforcement Learning Algorithm for Model-based Control

Xingxing Liang[1#], Qi Wang[2,1#], Yanghe Feng[1*#], Zhong Liu[1], Jincai Huang[1]

[1]College of Systems Engineering, University of Defense Technology, Changsha, China

[2]Institute for Advanced Study, University of Amsterdam, Amsterdam, The Netherlands

*Correspondence Author: fengyanghe@yeah.net

#These Authors Contributed Equally in this Work.

**Abstract**

Recent breakthroughs in Go play and strategic games have witnessed the great potential of reinforcement learning in intelligently scheduling in uncertain environment, but some bottlenecks are also encountered when we generalize this paradigm to universal complex tasks. Among them, the low efficiency of data utilization in model-free reinforcement algorithms is of great concern. In contrast, the model-based reinforcement learning algorithms can reveal underlying dynamics in learning environments and seldom suffer the data utilization problem. To address the problem, a model-based reinforcement learning algorithm with attention mechanism embedded is proposed as an extension of World Models in this paper. We learn the environment model through Mixture Density Network Recurrent Network(MDN-RNN) for agents to interact, with combinations of variational auto-encoder(VAE) and attention incorporated in state value estimates during the process of learning policy. In this way, agent can learn optimal policies through less interactions with actual environment, and final experiments demonstrate the effectiveness of our model in control problem.

# 1. Introduction

Reinforcement learning is a promising paradigm in complex scheduling and control tasks, and it has contributed a lot in domains, ranging from strategic game play [1, 2], unmanned aerial vehicle control [3], autonomous driving [4] to robots cooperation [5]. These remarkable achievements have not only demonstrated the plausibility and effectiveness of reinforcement learning for combating uncertainty from both environments and decision process but inspired the ever-lasting research interest in this domain as well [6]. One of the indispensable factors accounting for the success in applications is the agent's ability of successively interacting with environment. More specifically, the interactions relieve the extent of uncertainty, reveal the dynamics in environment and drive the agent to perform task-beneficial actions after learning from delayed rewards. Thus, such learning paradigm is also referred to as learning from error-trial signals and believed to pave the way to general artificial intelligence.

Roughly speaking, there exist two families in reinforcement learning. One is called as model-free based reinforcement learning, which merely takes advantage of rewards but neglects other inherent potential environment information, which would possibly promote the learning performance. With no dependence on implicit environment information suggesting properties of transitions or reward signals, model-free based learning enjoys its popularity in practice. However, a long-standing challenge for this type of learning is that millions of instances or experience need to be continually collected for policy evaluation and improvement. Meanwhile, the low efficiency of data utilization in model-free based learning is a waste of resource and restricts the power of reinforcement learning being applied to more universal real-life problems. On the contrary, another family, called as model-based reinforcement learning algorithms, does not strictly rely on sampling, enables characterization of potential dynamics in environment and mostly reveals the discipline of task. Once the model of Markov decision process(MDP) is approximately explored, there is no need to create additional experience through interactions with environment, and optimal policy can be directly derived based on the model. As evidenced in former works [7-9], model-based algorithms tend to maintain higher efficiency in some scenarios. Still, the bottleneck of model-based algorithms comes from weak capability in modifications and strongly dependence on precisely modeling, resulting in less adaptability and flexibility to dynamic Zero-shot or noisy environments.

As already noticed, model-free algorithms can achieve better performance, while model-based algorithms can well exploit dynamic properties of environment. Though some works have already tried to combine these two families in the past few decades, including synthetic experience generation [10, 11] and partially model-based backpropagation [7, 12, 13], the way to perfectly establish connections between two families remains limited. From the intuition, human's experience in life and work indicates the accommodation to complex environments may not demand so many instances to learn and summarize. Information we perceive every day with senses is quite limited, but human beings can easily generalize knowledge or skills regardless of diversity of scenarios. That is, we can conceptualize things from limited sensory information and generalize decision-making in various scenarios. One of possible reasons explaining this phenomenon can be the ability of abstraction, and once concepts as well as relationships between concepts are built with limited accessible data, we can establish some abstract model to represent the environment and partially predict dynamic variations of environment given some actions, including state transitions and reward signals. Similar interesting viewpoints refer to existing works [14, 15] on neural network models, insisting human beings tend to build world physical model with finite cognitions, and perform decision-makings based on mentally constructed model. Instead of learning new models, our brain can make decisions more frequently with the formerly self-constructed physical model in our mind [16, 17]. By predicting the future scenario after instant action at some state, we can promptly react and avoid potential danger with previously built model [18, 19]. A recent trial can be seen in World Model(WM) [15], and D. Ha & J. Schmidhuber [15] has put it in practice, in which environment model is built with limited real world experiences and policy learning efficiency is proved to be advanced. One of significant benefits of such framework in modeling is that heavy workloads of collecting transitions in environment as well as concerning expense would be brought in reduction through learning a virtual environment model. For an instance, unimaginable massive images in a variety of real scenarios iteratively perceived by sensors are required to feed an autonomous driving system, and the generalization capability of such system is theoretically positively correlated to manual and financial expense in accessing environmental data. In some sense, these consuming prediction or control tasks can be partially addressed through interactions with well-trained virtual environment. In this paper, we make an extension of the former work [15] and study the approach to aggregating both model-free algorithm and model-based control to explore abundant environment information buried in experience and further guide the optimal policy search. More technically, we also make use of neural networks in state embedding representations, sequence data prediction and improve WM through attention-based policy learning. The remainders are arranged as follows. Section 2 summarizes some related works and express our intentions in research. Basic knowledge about reinforcement learning is included in Section 3, which would contribute to optimization process in our model. Section 4 would pay a revisit to WM and elaborate framework of our proposed model, including components in the model, training procedures and technical details. Section 5 is about the experiments in some classical control problem, and we also analyze the performance and sensitivity to parameters. Finally, some conclusions are drawn and some future works with respect to this domain are highlighted.

## 2. Literature Review

The simulation environment is helpful in developing and testing new reinforcement learning algorithms, and OpenAI Gym [20] provides a series of virtual environments to carry out experiments, allowing the comparison and validation of algorithm performance. These include some traditional problems of control, among which end-to-end tasks are more practical but challenging. The end-to-end tasks push the agent to directly receive original input such as images of some scenarios as signals to make decisions, including Cart-Pole control and Car-Racing. The inherent high dimensionality of images poses great difficulty in learning process and inspires the application of representation learning to reinforcement learning [21]. Encoding a complicated instance into a vector of low dimensions, deep neural networks can extract compact representations for the input of high dimensions, including but not limited to images. The representation learning with deep neural networks makes it possible to train a reinforcement learning model in dealing with complicated tasks. Another advantage of using deep learning lies in better generalization, and both of DQN [1] and AlphaGo Zero [2] have benefited from the convolutional neural network's representations of states and achieved state of art performance in policy learning.

Though powerful representation model and increasingly computational power can satisfy the basic demands of solving complicated control problems with reinforcement learning, the access to the dataset from real environment is still the bottleneck in this domain and the algorithm is hungry for this resource in some sense. Honestly speaking, interactions with environment are decisive to the successful application of reinforcement learning, and to achieve ideal performance, a wealth of resources such as human labor, time and money are consumed to collect transitions and rewards from environments. Especially for model-free based reinforcement learning algorithms, the circumstance is more obvious, and the lower efficiency of data utilization accompanied with neglect of structure information in environment is another urgent concern. Such dilemma has caught increasingly attention in domains and inspired some interesting thoughts to tackle problems.

Learning the environment is extremely crucial in this study, and there exist mainly two paradigms for capturing the properties of environments and relieving bias in modeling. One is to learn samples indicating properties of generated environment in the form of some probability distribution and explore the policy as well. Earlier works on simultaneously learning environment model and policy are not stable, while expectation maximization(EM) [22] can separately capture the environment model, disentangle the parameters from control model and just learn limited control parameters to accelerate the rate of convergence. As a breakthrough in learning environment model, WM [15] can automatically reveal dynamics environments and its motivation from cognition science has been mentioned in Introduction [15]. A. Piergiovanni et al. [23] constructed deep neural networks to encode states and predict future scenario as a simulation of environment model, and demonstrated the robot can learn plausible policy to act in real world through interacting with such dreaming environment. Noticing the high complexity and cost to handle image observations in visual based reinforcement learning [24], A. V. Nair et al. developed a reinforcement learning algorithm with imaged goals, which combined variational auto-encoder(VAE) with off-policy goal-conditioned reinforcement learning. To address planning problem in uncertain environments, a recurrent state space model was trained to capture dynamics of environment in pixel level and the constructed agent called Deep Planning Network can learn policies to control [25]. Additionally, the original image is seldom used in environment modelling, e.g. World Models [15] and PA [26], and auto-encoder is mostly introduced to represent the state in low dimensions, further advancing the training efficiency and bringing reductions on the scale of parameters in control. And another paradigm is motivated by meta-learning, which seeks multiple dynamic models learned from various environments and integrates the characteristics of these models to describe the uncertainty in environment [22, 27, 28].

## 3. Background

Reinforcement learning is a traditional learning paradigm to deal with prediction and control problems in uncertain environments. The main goal of reinforcement learning is to capture some policy to maximize the cumulative rewards, which means selecting proper action given some states. Generally, it can be described with Markov Decision Process(MDP), which is formulated in a tuple of five elements $\{S, A, R, P, \gamma\}$. And elements in tuple respectively represent the set of states in environment $S = \{s^{(i)} | i = 1,2,..n\}$, the set of available actions in environment $A = \{a^{(i)} | i = 1,2,..,m\}$, the reward function conditioned on state transition in environment and some action $R = \{r(s_{t+1} = s | s_t = s', a_t = a) | a \in A; s, s' \in S\}$, the transition probability between states given some action $P = \{p(s_{t+1} = s | s_t = s', a_t = a) | a \in A; s, s' \in S\}$ and the discount of step reward in long-run experience $\gamma$.

Mathematically, the cumulative rewards with discount factor $\gamma$ and initial state s under some policy $\pi$ is $R = E_\pi[\Sigma_{t=0}^{T} \gamma^t r_t | s_0 = s]$, where $\{r_t | t = 0,1,2..,T\}$ is the set of reward signals in each time of state transition after some action.

And the policy to learn is a map from state space to action space as
$$\pi: S * A \to [0,1]$$
$$\pi(a|s) = p(a_t = a | s_t = s)$$

The learning process is to interact with environment, collect some experiences with state transitions and rewards information and evaluate and improve policies.

## Proximal Policy Optimization(PPO)

In cases of nonconvex optimization, gradient can be computed with numerical or sampling methods, but a propriate learning rate in iterations is hard to determine and it need to vary with time to ensure better performance. Earlier works on reinforcement learning also encounter such dilemma when using gradient based optimization technique, and simulated annealing algorithm is widely used to determine learning rate with annealing factor during optimization process, gradually decreasing the step width of learning rate. However, it is still tough for policy gradient-based reinforcement learning algorithms to modify learning rate especially when training neural network.

To circumvent the bottleneck, Schulman et al. [29] proposed Trust Region Policy Optimization(TRPO) algorithm to deal with random policy, in which Kullback-Leibler(KL) divergence between old policy and updated policy is considered in objective function, and the KL divergence in each state point can be bounded as well. The approach jumps out of modifying learning rate, enforces the process of policy improvements more stable and is theoretically proved to monotonically increase the cumulative rewards. Considering the complexity of second order Hessian matrix computations in TRPO, Schulman et al. [30] further developed one order derivative proximal policy optimization(PPO) algorithm.

The surrogate loss function in original TRPO can be formulated as

$$\max_\theta L^{CPI}(\theta) = \max E_t^\sim [r_t(\theta) A_t^\sim], r_t(\theta) = \frac{\pi_\theta(a_t|s_t)}{\pi_{\theta_{old}}(a_t|s_t)} \;.$$

Where $\pi$ is some stochastic policy, $\pi_{\theta_{old}}$ is the parameters in policy in last time, and $A_t^\sim$ estimates the advantage function of performing $a_t$ conditioned on the state $s_t$ at time step t. The objective as the expectation is the empirical average over instances in mini-batch.

Through pruning the above surrogate loss function, we can obtain loss function in PPO as

$$L^{CLP}(\theta) = E_t^\sim [\min(r_t(\theta) A_t^\sim, clip(r_t(\theta), 1-\epsilon, 1+\epsilon) A_t^\sim].$$

Where the function *clip* is

$$clip(x, x_{MIN}, x_{MAX}) = \begin{cases} x, if\ x_{MIN} \leq x \leq x_{MAX} \\ x_{MIN}, if\ x < x_{MIN} \\ x_{MAX}, if\ x > x_{MAX} \end{cases}$$

And the objective can be optimized with stochastic gradient descent, minimizing the KL divergence and reducing workloads of modification.

## 4. Methodology

In the former sections, some basic knowledge about reinforcement learning have been introduced, and concerning challenges are presented. In this section, we would elaborate our proposed model, referred to as VMAV-C, in decision-making process. Here, VMAV-C corresponds to a combination of **V**ariational Auto-Encoder(VAE), **M**ixture Density Network-Recurrent Neural Network(MDN-RNN), **A**ttention-based **V**alue Function(AVF) and **C**ontroller Model. Different from the covariation matrix adaptation evolution strategy utilized to optimize Controller Model in WM [15], we make use of PPO based Actor Critic(AC) algorithm [31] in **C**ontroller, and attention mechanism is considered in critic network for estimates of state value function. In AC algorithm, critic network generally works in prior to actor network, since precise estimation of value function can better accelerate the policy learning.

At first, some fundamental components in original framework VM-C [15], including VAE, MDN-RNN and Controller, are detailed as the background. And then the attention-based value function is highlighted, and the way to combine with Critic network is core of the framework. To understand how VMAV-C works, training process would be separately discussed.

## 4.1 Outline of VM-C Model

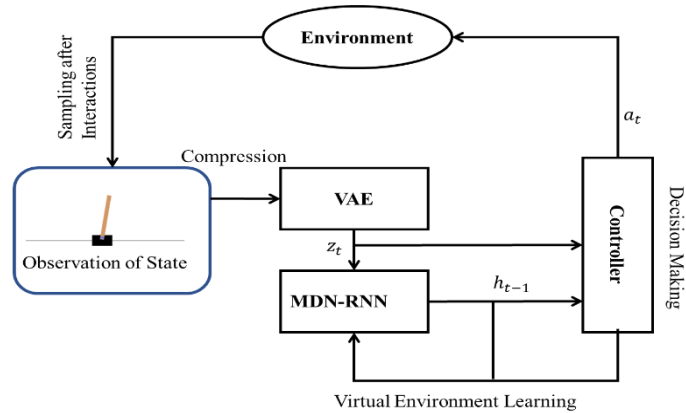

**Fig1. Framework of VM-C used in World Models.**

The **Fig1** reveals the relationship between VAE, MDN-RNN and Controller model, and answers the question on how VMC dynamically reacts to the environment. The lines in **Fig1** indicate the information flow and control operations in given environment. The specific procedure in platform of OpenAI gym can be described as follows.

---

**Decision Process with VMC：**
obs=env.reset()
h=rnn.initial_state() //initialize Recurrent Neural Networks
done=Flase
cumulative_reward=0 //initialize the cumulative rewards when interacting with environment
While not done:
    z=vae.encode(obs) //encode the observation of state in latent representation
    a=controller(z,h) //input latent representation of observation and hidden information in //RNN
    next_obs, reward, done, _=env.step(a) //perform action and receive responses
    cumulative_reward +=reward //cumulative reward in each step
    h=rnn.forward(a, z, h) //compute next time hidden information in RNN
    obs=next_obs
return cumulative_reward

---

### 4.1.1 VAE

Formulated as the information compression technique, the auto-encoder (AE) attempts to encode the original input into a vector of fixed length and then to decode such latent representation to reconstruct the input. As one of the commonly used AEs, variational auto-encoder (VAE) is used in learning the latent representation of some complex datasets or manifolds through variational approximations and reproducing some synthetic instances through sampling from latent space[32]. The ideology of VAE is that some complicated dataset can be probabilistically generated from some latent variables through a series of transformations. To enable the computation process tractable and efficient, VAE assumes variables z in latent space obey some multi-dimensional Gaussian distribution $z \sim N(0, I)$. Though the distribution of latent space can be more universal according to specific hypothesis, the continuity of Gaussian distribution and the differentiation of neural network connections makes it viable to utilize the back propagation.

More specifically, given some instance x, we wish to uncover its relationship with latent variables z, so some neural network $q(z|x)$ as encoder is used to approximate actual latent variable distribution $p(z|x)$. With the help of Bayes theorem, the Kullback-Leibler divergence is computed as the difference between two distributions

$$D_{KL}(q(z|x)||p(z|x)) = E_{z \sim q(z|x)} \log q(z|x) - E_{z \sim q(z|x)} \log p(z|x) = E_{z \sim q(z|x)} \log q(z|x) - E_{z \sim q(z|x)} \log p(x|z) - E_{z \sim q(z|x)} \log p(z) + \log p(x).$$

Equivalently,

$$\log p(x) = E_{z \sim q(z|x)} \log p(x|z) - D_{KL}(q(z|x)||p(z)) + D_{KL}(q(z|x)||p(z|x)) \geq E_{z \sim q(z|x)} \log p(x|z) - D_{KL}(q(z|x)||p(z)).$$

The left-hand side of inequality, which we wish to maximize, is computationally intractable, so the right-hand side term also referred to as evidence lower bound(ELBO) is taken as the objective, and the first term of ELBO is decoder in form of neural network structure. Hence, VAE makes a trade-off between two loss functions, respectively as the construction error in decoder and the KL divergence in encoder, and conceptually connects observed space to latent space probabilistically with some differentiable neural network.

Recognized as a typical generative model, once VAE is well trained through back propagation, it can draw some samples of compressed representations from latent space distribution $p(z)$ and then generate novel instances through decoder network $p(x|z)$.

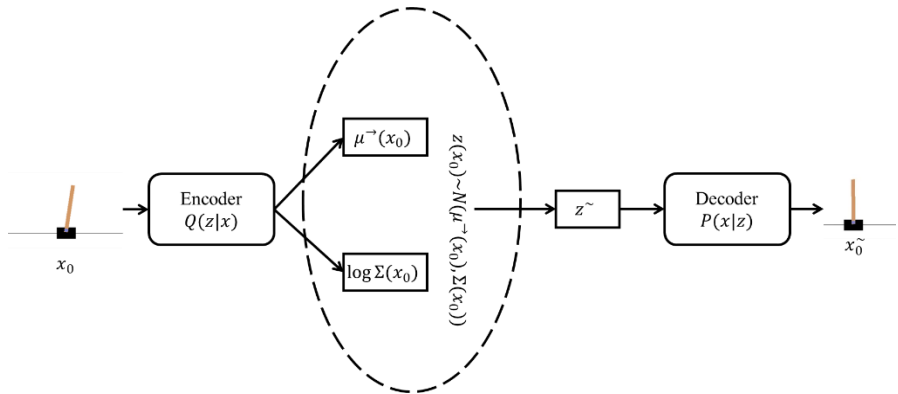

**Fig2. VAE in Observation of CartPole-V0.** Encoder and Decoder are two neural networks, and mean vector and logarithm variance vector are latent representation for some state.

As displayed in **Fig2**, the input of VAE in our experiments is the observation of environment, namely scenario image of CartPole-V0, and we compress this observation into some low dimension vector as the latent representation.

**4.1.2 MDN-RNN**

With the mixture density model and conventional neural network aggregated, mixture density network can approximate arbitrary conditional probability distributions, especially those with continuous input, and solve the inverse problems in practice. Meanwhile, recurrent neural network(RNN) is proved to be efficient in capturing the dependencies in sequence datasets and perceiving the trend of sequence in some sense. Hence, some works focus on the combination of these two techniques and present some variants of RNN, referred to as MDN-RNNs, in dealing with real life problems [15, 33], and a recent interesting research is about applying MDN-RNN to sketch generation in drawings [33].

For general purpose, RNN is used to model the conditional probability distribution $p(z_{t+1}|a_t, z_t, h_t)$, where $a_t, z_t, h_t$ respectively correspond to action, latent representation of state and sequence hidden information at the time step t, and $z_{t+1}$ is the predicted state in time step $t + 1$ conditioned on $a_t, z_t, h_t$. When RNN meets reinforcement learning tasks, some modifications are required for probability representations, since information whether the episode ends in process need to be marked. Thus, with the involvement of additional variable $d_{t+1}$ to indicate whether the episode ends at time step $t + 1$, the probability can be parameterized in the form of $p(z_{t+1}, d_{t+1}|a_t, z_t, h_t)$, which has been illustrated in **Fig3**. Especially, $d_{t+1}$ is a binary variable to predict, and the episode is predicted to end when its value turns to one instead of zero.

Gaussian mixture model(GMM) is a mixture probability model, and take the bivariate response variable as an example, GMM with m components of Gaussian distribution can be linearly represented as follows:

$$p(x, y) = \Sigma_{j=1}^{M} \theta_j N(x, y | \mu_{x,j}, \mu_{y,j}, \sigma_{x,j}, \sigma_{y,j}, \rho_{xy,j})$$

where $\Sigma_{j=1}^{M} \theta_j = 1$ and $\{\mu_{x,j}, \mu_{y,j}, \sigma_{x,j}, \sigma_{y,j}, \rho_{xy,j}\}$ is respectively means, standard variances and correlation coefficients for Gaussian distribution indexed with j. In the mixture density network, exponential function, hyperbolic tangent function and softmax function are respectively employed to normalize the variance, correlation coefficient, and prior weight $\sigma = \exp \tilde{\sigma}$, $\rho = \tanh \tilde{\rho}$, $\theta_k = \frac{\exp(\tilde{\theta_k})}{\Sigma_{j=1}^{M} \exp(\tilde{\theta_j})}$. To control the randomness of sampling in Gaussian distribution, the temperature parameter τ is used to adjust the scale of prior weights and variances $\tilde{\theta_k} \to \frac{\tilde{\theta_k}}{\tau}, \sigma^2 \to \tau \sigma^2$.

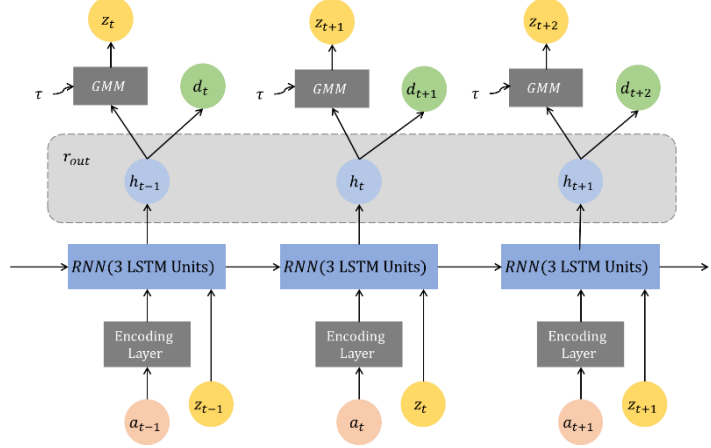

**Fig3. MDN-RNN.** Each Box of LSTM Networks contains three LSTM units.

In our learning task, we also encode the discrete action as $f(a_t)$[1] and combine it with latent representation of state $z_t$ and hidden information $h_t$ at time step t to guide the prediction of future state in environment as $p(z_{t+1}, d_{t+1}|f(a_t), z_t, h_t)$. Additionally, the loss function of MDN-RNN comprises two components: the prediction error in next state $L_s$ and the prediction error in mark of ending state $L_p$.

$$L_s = -\frac{1}{N}\Sigma_{i=1}^{N} \log(\Sigma_{j=1}^{M} \theta_j N(x,y|\mu_{x,j}, \mu_{y,j}, \sigma_{x,j}, \sigma_{y,j}, \rho_{xy,j}))$$

$$L_p = -\frac{1}{N}\Sigma_{i=1}^{N}(\alpha d_{(t+1)_i} \log q_i + (1 - d_{(t+1)_i}) \log(1 - q_i))),$$

where $1 - q_i$ is the predicted probability when mini-sequence ends at time step i.

As the proportion of ending states is quite limited, we place more weights of penalty on these instances through enlarging the value of $\alpha > 1$ in $L_p$.

Finally, the total loss function is the weighted sum of two loss functions:

$$L_{total} = \beta_1 * L_s + \beta_2 * L_p,$$

where $\{\beta_1, \beta_2\}$ is the set of weights in loss terms.

**Fig3** details the structure of MDN-RNN in our model and indicates dependencies between action, latent representation of state, hidden information in mini-sequences of epoch and ending state of mini-sequence.

### 4.1.3 Controller Model

Functioned as a decision maker, Controller Model in **Fig4** plays a critical role and is expected to seek optimal action given specific state in each time step to maximize cumulative rewards. That is, MDN-RNN produces former well-encoded hidden state information as well as current state information for Controller, and the latter determines which action to select as $a_t \sim \pi(a|z_t, h_t)$. Specifically, the former mentioned state-of-art algorithm PPO is taken in our paper for policy learning.

---

[1] In our task, discrete action is initialized as one-hot encoding and through a two-layer fully connected neural network a 32-dimension vector is learned as the $f(a_t)$.

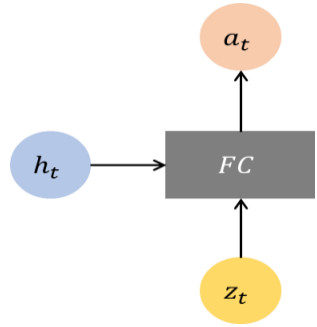

**Fig4. Controller Model.** $h_t$ comes from hidden information in MDN-RNN, $z_t$ is latent representation of current state, and action is conditioned on both information. Fully connected(FC) network is used here.

## 4.2 Involvement of AVF

Recently, attention mechanism is attracting increasingly attention and frequently accompanied with sequence learning, due to its fancy power in performance promotion in comparison to ordinary sequence models. The operation of attention mechanism is to probabilistically assign various weights to hidden information of historical sequence and then aggregate them to form context vector for some predicting time steps, and the hidden information more related to predicting time step would receive more attention and be assigned more weight. That is, given hidden information of some t-length sequence $H = [h_1, h_2, .., h_t]$, the context vector v for predicting time step serves as the embedding information for historical sequence and is computed as the weighted sum of hidden information in such time step $v = \Sigma_{i=1}^{t} \alpha_i h_i$.

During the training process with reinforcement learning algorithm, we incorporate the attention mechanism in estimation of state value function, which suggests historical hidden information may contribute to the estimation of current state value in varying weights.

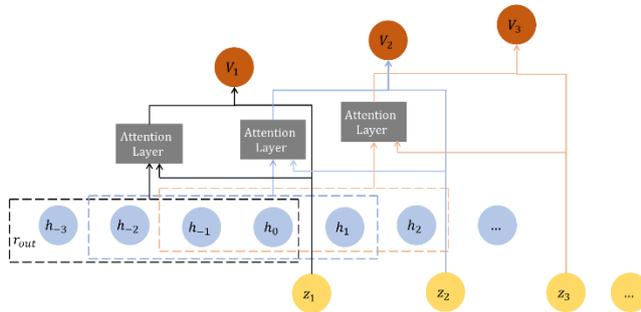

**Fig5. Attention-based Value Function Representation.** Four recent units containing hidden information from MDN-RNN contribute to state value estimation, and attention layer is to compute importance of these information.

In Critic network of AC algorithm, hidden information of each time illustrated in **Fig5** comes from MDN-RNN $r_{out}$ layer. And historical information in former n time steps is utilized for current state value estimation. To ensure the initial state can also satisfy structure of attention, absent previous hidden information such as $\{h_{-3}, h_{-2}, h_{-1}\}$ in $[h_{-3}, h_{-2}, h_{-1}, h_0]$ is initialized with zeros (Here take the case in **Fig5** as an example).

Thus, the context vector with attention can be computed as

$$c_t = \Sigma_{i=1}^n \alpha_i h_{t-i}$$

$$\alpha_i = \frac{\exp(\widetilde{\alpha_i})}{\Sigma_{j=1}^n \exp(\widetilde{\alpha_j})}$$

$$\widetilde{\alpha_i} = W[h_{t-i}, z_t] + b$$

Where $w, b_i$ are the parameter in RNNs and $\alpha_i$ reflect the strength of impact of historical information indexed with i in context vector $c_t$. $z_t$ as the input in predicting time step is the next step of state information. For the state value function estimation, it is required to combine both context information $c_t$ derived from previously hidden information $\{h_{t-1}, h_{t-2}, .., h_{t-n}\}$ and current state information as

$$V(s_t) = W_v[z_t, c_t] + b_v.$$

Where $z_t$ is the latent representation of state in time step t, $c_t$ is the context vector with attention, $[.,.]$ is the concatenation of vectors and $\{W, b, W_v, b_v\}$ is the set of parameters to learn in attention-based value neural network. **Fig5** reveals the learning process of state value, and this structure is also the specific setting in our experiments.

## 4.3 Training Details for VMAV-C RL

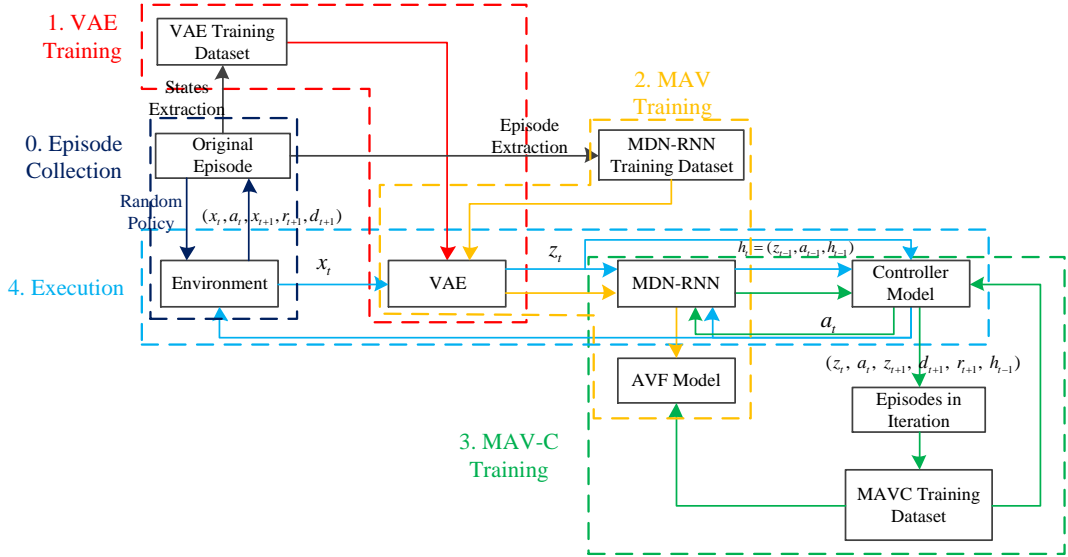

**Fig6. VMAV-C Reinforcement Learning Training Framework.** Arrows suggest information flow in modules.

The observations, which agent receives or recognizes from actual environment are raw 2-D images with high dimensions, and the role of VAE in experiments is to compress images in lower dimensions. As illustrated in **Fig2**, VAE captures the latent representation of observation from the hidden layer and reconstructs the observation. Note that this preprocessing of states can be trained independently from the whole model training.

Since the latent representations of observations in environment are in time series with the episode going, the correlation information as well as transition characteristics in environment is buried in the latent space as well. MDN-RNN makes attempt to generate latent representation of observation $z_{t+1}$ and predict the ending state of sequence $d_{t+1}$ in next time step conditioned on current state $z_t$ and hidden sequence information $h_t$ after some specific action $a_t$. Of course, the predicted latent representation of observation can be decoded for visualization through decoder layers of VAE to understand the next observation. In the light of uncertainty in transitions of environment, the probability function $p(z)$ is utilized to estimate possible embedded representations of states in the future. As formerly mentioned, GMM is trained to approximate the probability distribution of embedded representation of state information of environment in next time as $p(z_{t+1}, d_{t+1}|a_t, z_t, h_t)$. Similar to the work [33], the temperature parameter $\tau$ is also included in sampling process to combat uncertainty. Intuitively, MDN-RNN would learn a model simulating actual environment, which suggests how the environment varies with time conditioned on some state after performing some specific action. And in our model, MDN-RNN would be pretrained as initialization for training process.

The Controller Model directly takes part in in training process and execution process. The reason why we introduce attention in Critic network in Controller model is due to different importance of historical hidden information in sequence for state value function estimation, and the following experiments would demonstrate varying weights of importance on former sequences do bring better generalization.

More concrete illustration on training process is in **Fig6**, and the formerly induced components or subset are tied in five modules, respectively as Episode Collection(Step 0), VAE Training(Step 1), MAV Training(Step 2), MAV-C Training(Step 3) and Execution(Step 4). And dependencies between components in different modules are also summarized in dotted boxes in **Fig6**. Notice that these steps are in order of training and testing model, and detail roles of modules are described in following sections.

### 4.3.1 Pretraining Details

The purpose of VAE, MDN-RNN, AVF and Controller is to learn representations of states and the dynamic transitions in the environment at the same time, but massive parameters and complexity of network structures make it tough and time-consuming to train VMAV-C. Hence, synchronously pretraining VMAV is the required step in our experiments. To achieve this aim, we collect 2000 episodes with random policy strategy through a series of interactions with actual environment as $\{episode = \{(x_t, a_t, x_{t+1}, r_{t+1}, d_{t+1})\}\}$ in **Step 0**. These rollouts/screenshots of the environment serve as the training dataset for VAE, and we assume the sampling has approximately encompassed the dynamic information of the environment, especially the state representations and concerning transitions.

Further in **Step 1**, the whole dataset of states as the input for VAE are randomly partitioned into two sections, and 75% are for the training process while the rest are for testing the reconstruction performance. During this process, the latent space for environment in the form of image frames is efficiently explored by monitoring the reconstruction error in testing dataset. Here, the images are reduced into 64 dimensions in latent space, since the dimension of multivariate normal distribution of latent space $z$ is assumed as 32. Once the training process of VAE is completed, the collected images can be encoded in lower dimensional vectors as partial inputs for MDN-RNN model.

After that, we can access to latent representations of states in formerly collected episodes, which are embedded with VAE. And these episodes are firstly merged into one long sequence according to order of time and then sliced into mini-sequences of fixed length as dataset to learn MDN-RNN. In our experiments, the length of each mini-sequence is 32 steps in state transitions, and the mini-sequence would include some time step when epoch ends. After a few iterations as initialization for MDN-RNN, AVF is also simultaneously included in pre-training process. Some modification on ending time step is performed during AVF training as the random initialization of historical hidden information $h_t$. In this way, AVF is preliminarily learned through pre-training, and a virtual environment buried in MDN-RNN is derived. These are the goals in **Step 2** to achieve.

**Algorithm1. Pretraining VMAV-C Model**

Input: Initialized VAE, MDN-RNN, Controller and Attention Value Model initialized with random policy

Output: Trained VAE, Pretrained MDN-RNN, Pretrained Attention Value Model

(1) Rollout to actual environment N times with random policy. Save all actions, observations, rewards and ending indicators $\{episode = \{(x_t, a_t, x_{t+1}, r_{t+1}, d_{t+1})\}\}$ during rollouts to storage device

(2) Collect observations of states $\{x_t\}$ to train VAE

　　**While** VAE has not converged **Do**:

　　　　Sample mini-batch of states in images

$$loss_{vae} = \frac{1}{N}\Sigma_{i=1}^{N}[(VAE(x_i) - x_i)^2 + \frac{1}{2}\Sigma_{j=1}^{k}(\mu^{(j)^2}_{xi} + \sigma^{(j)^2}_{xi} - \ln\sigma^{(j)^2}_{xi} - k) \qquad //\text{The}$$

　　　　//number of examples in mini-batch N, the dimension variable z as latent space in VAE k

　　　　Update VAE through Back Propagation // RMSProp as the default Optimizer

(3) Collect MDN-RNN training dataset

　　**For** episode in storage:

　　　　Transform the episode to some fixed length sequence with L time steps

　　　　**For** each time step:

　　　　　　Formulate transition as $(z_t = VAE_{Enco}(x_t), a_t, z_{t+1} = VAE_{Enco}(x_{t+1}), r_{t+1}, d_{t+1})$

　　　　Store these mini-sequences of time steps in memory as $M_{MDN-RNN}$

(4) Train MDN-RNN

　　**While** MDN-RNN has not converged **Do**:

　　　　Sample batch from $M_{MDN-RNN}$

　　　　Compute total loss function $L_{total} = \beta_1 * L_s + \beta_2 * L_p$ in **Section 4.1.2**

　　　　Update MDN-RNN through Back Propagation //Adam as the default Optimizer

(5) Train AVF in AC algorithm

　　**While** AVF has not converged **Do**:

　　　　Sample mini-batch from $M_{MDN-RNN}$ in step (3)

　　　　Generate $r_{out}$ and formulate training dataset as **Fig5**

　　　　Use n-step return:

$$\tilde{V} = \begin{cases} \Sigma_{t=0}^{T-1}\lambda^t r + AVF(z_T, h_T), if\ d_T = 0 \\ \Sigma_{t=0}^{T-1}\lambda^t r + 0, if\ d_T = 1 \end{cases}$$

$$loss_{AVF} = -E(\tilde{V} - AVF(z,h))^2 \quad //\text{T is the number of maximum of steps return}$$

//rewards, here 32 is the maximal for T because of length of mini-sequences.
Update AVF through Back Propagation //Adam as the default Optimizer

Notation: Here $\{\beta_1, \beta_2\}$ is the set of hyper parameters to adjust the learning goals in different stages.

### 4.3.2 Training Details

In **Step 2**, we have learned a virtual environment model based on MDN-RNN, which can theoretically reveal transitions of states and reward signals. Hence, the following procedure would be running reinforcement learning in such environment. In **Step 3**, AVF and Controller are trained through interactions with the virtual environment, and we utilize PPO algorithm to optimize Controller Model. We expect the virtual environment learned in pre-training as MDN-RNN has included dynamical properties of environment as precisely as possible and can function as the actual environment here.

In **Step 4**, VAE, MDN-RNN and well-trained Controller from **Step 3** are used in making decisions in the actual environment, since Controller is conditioned on information from latent state representation z produced by VAE and historical hidden information h produced by MDN-RNN. Notice that apart from the use of the virtual environment information in policy learning, **Step 4** in some sense utilizes the actual environment information from sequential reward signals to improve policy as well.

**Algorithm2. MVA-C Model Training with PPO**

Input: Trained VAE, Well-trained MDN-RNN, Well-trained Attention Value Model, and Controller Model

(1) Environment initialization to obtain initial state
(2) **For** i = 0,1,.., K :
   Drive the agent to interact with virtual environment (MDN-RNN in pre-training process) to collect hidden information of RNN h, latent representations z embedded with VAE, action a and feedbacked reward signal r.
   Optimize the policy through PPO:
   $$L^{CLP}(\theta) = E_t^{\sim}[\min(r_t(\theta)A_t^{\sim}, clip(r_t(\theta), 1 - \epsilon, 1 + \epsilon) A_t^{\sim}]$$
   Update state value network with minimization of mean square error(MSE):
   $$\min_w E(\Sigma_{t=1}^T \lambda^{t-1} r_t + AVF(h_T, z_T, w) − AVF(h, z), w)^2 \quad //\text{w is the State Value Network}$$
   //Parameter
(3) End For

## 5. Experiments and Performance Analysis

In this section, we would carry out experiments to accomplish a representative end-to-end reinforcement task, CartPole-V0 control based on the OpenAI Gym platform [20]. In this background, a cart is attached and un-actuated joined with a pole. The control decisions include an implementation of a force of +1 or -1 to the car. The pendulum is initialized upright, and the prolonging the time of keep upright during the process is the target in the task. The end of the episode is when the cart moves away from the ranges of 2.4 units in the center or the pole is over the 15 degrees from the vertical. For the reward, the +1 signal is returned every timestep in the condition the pole remains upright.

Based on this end-to-end control problem, we would carry out a series of experiments with VMAV-C to test performance.

## 5.1 Baseline Algorithms

To demonstrate the effectiveness of our proposed model VMAV-C, two baseline algorithms are introduced for comparison. They are respectively Contractive PPO(CP) algorithm and MDN-RNN PPO(MRP) algorithm.

1. Contractive PPO algorithm. This algorithm receives raw images from actual environment and throw them into VAE to obtain encoded representations as inputs for agent to make decisions. And the agent directly interacts with the actual environment, and Actor Critic algorithm with PPO is also used in learning policy. That is, the process of learning model of virtual environment is not required here.
2. MDN-RNN PPO algorithm. This algorithm is a variant of method in Ha D et al.'s work[15], but the PPO is used in Controller model to learn policies instead of evolutionary algorithms. And the main difference from our model is the lack of attention mechanism in AC.

The settings are detailed in **Section 5.2**, and the performance of running three algorithms are already summarized in **Section 5.3**.

Additionally, the scenario when the agent is completely trained to make decisions in virtual environment is also included in these experiments, which is specifically highlighted in **Section 5.4**. That is, we would randomly sample some state from actual environment as the initial, run MDN-RNN in virtual environment, automatically generate future states given actions, feedback rewards and improve policy. And such learned policy is used in the actual environment to test the accumulative rewards without discount. This operation is different from **Step 4**, since the actual environment does not offer sequential reward signals in policy improvement but is merely for testing policy learned from virtual environment every fixed number of epochs. So Controller as the agent directly learns the policy through interactions with the virtual environment, and PPO is used for policy learning.

## 5.2 Experimental Settings

The original image size in environment of CartPole-V0 is 400*600, and we reduce the size into 160*320 at the center of cart position and further resize them into 40*80. To formulate dataset of environment states, we run 2000 episodes and record raw images of states, instant action, next frame of image, related reward and indicator of whether the episode ends through successively interacting with actual environment. The concrete structure of VAE in our model is demonstrated in **Fig7**, and the training process refers to Section **4.3.1** in **Algorithm1**.

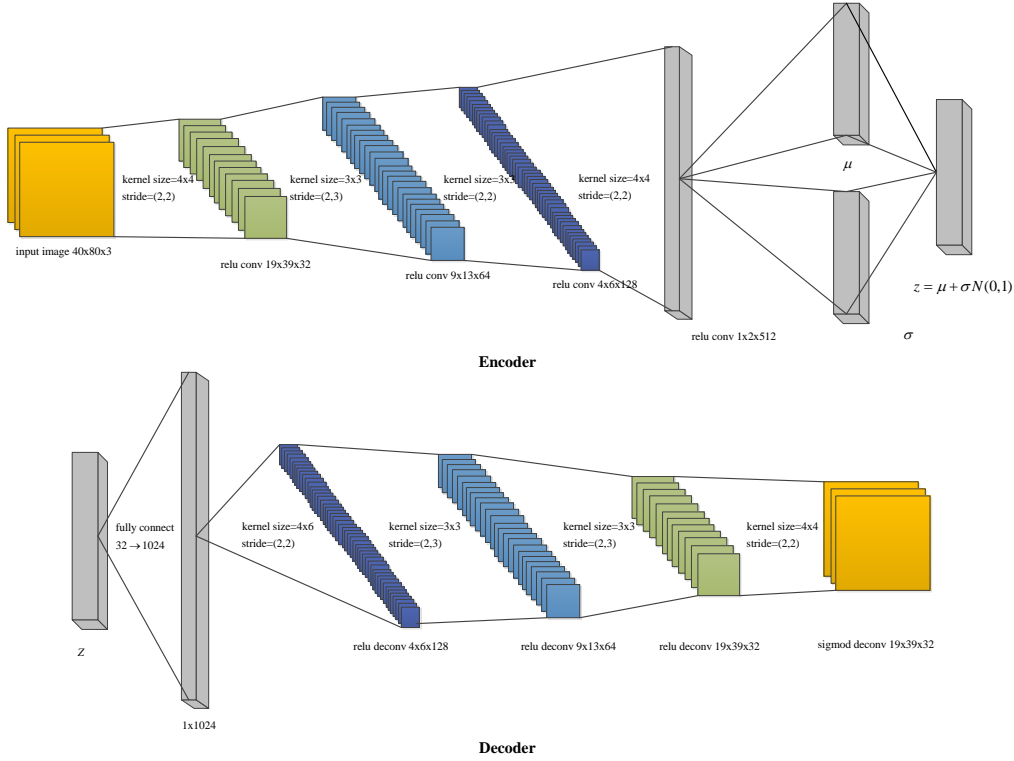

**Fig7. The Network Structure of VAE in our Experiments.** In experiments, the latent variable z obeys 32 dimensional multivariate normal distribution.

Once VAE is trained, we make use of it to encode the whole of formerly collected dataset of states into latent representations, which are employed as partial inputs for MDN-RNN training. And then about 1500 episodes of experience are merged as the training dataset for MDN-RNN, and we in order slice these into several mini-sequences to feed MDN-RNN at the step width 32. The rest of dataset, containing 500 episodes, is sliced in the same way to formulate testing dataset. However, this manipulation also brings some uncertainty about context information, since the context information for initial state in mini-sequence theoretically depends on last mini-sequence but the model fails to include. In the pretraining process, the parameters for MDN-RNN are set as $\beta_1 = \beta_2 = 1$, and the mini-batch size is 256, and Adam is chosen as the default optimizer with learning rate 4.77e-5, and the parameter $\tau$ to control randomness is 1. For the loss $L_p$ in total loss of MDN-RNN, parameter $\alpha$ is tuned as 2 for mini-sequences with ending mark of state. Through a network of three layers long short term memory(LSTM) units, MDN-RNN encodes the context information of mini-sequence into vectors and then learns the means and logarithm variances of five Gaussian distribution as well as corresponding prior weights in GMMs to make predictions towards next state of latent representation $z_{t+1}$ ending state $d_{t+1}$.

## 5.3 Policy Learning in Actual Environment

In the CartPole-V0 control problem, the mentioned three models, CP Model, MRP Model and VMAV-C Model, are employed and compared in performance. We would respectively analyze accumulative rewards without discount and loss of state value network separately and uncover potential reasons behind phenomenon in **Fig8**.

The number of iterations in training for these three models is 60000 and each iteration means the a sequence of up to 32 step transitions in interactions, and shadow curves which fluctuate fiercely are real cumulative values in results. To better display the results, Tensorboard [34] is used to smoothen these results into dark colored curves.

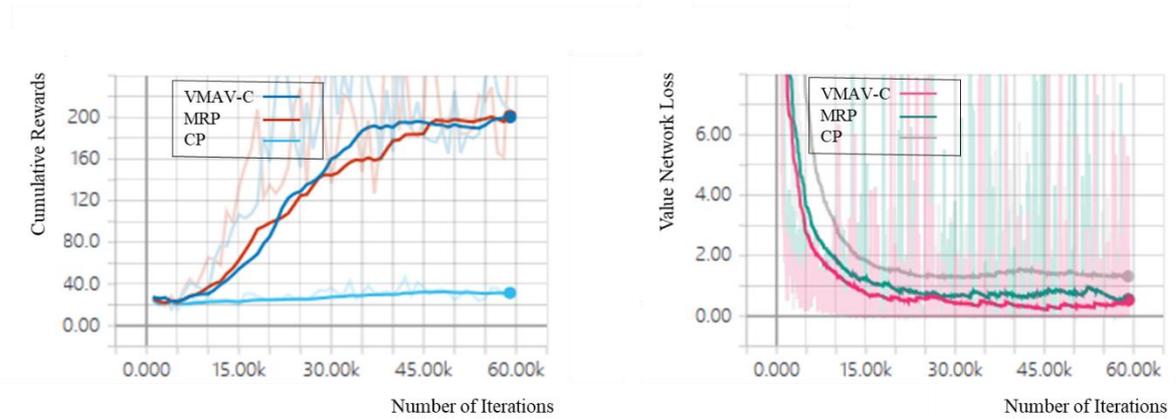

**Fig8. Cumulative Rewards and Value Network Losses with Three Models in the Actual Environment.** Smooth rates in test rewards and value network losses are respectively set as 0.9 and 0.95 in Tensorboard.

## Learning with CP Model

As mentioned before, CP Model also makes use of VAE to preprocess states and takes latent representations of original images as the input. And the model directly interacts with the actual environments.

From **Fig8**, we can notice that with value network loss function deceasing, agent can gradually improve policies, but this process requires massive experience and the learning speed is not ideal as well.

## Learning with MRP Model

Apart from state representations in low dimensions with VAE, MDN-RNN can learn the environment model to perceive possible transitions and rewards in the future. It can be observed in **Fig8** that MRP Model benefits from additional information provided by the learned environment model, and the required volume of experiences is smaller but with more satisfying rewards in comparison to CP Model. Meanwhile, there is a significant sharp declining tendency in value network losses.

## Learning with Attention-based World Model

In **Fig8**, we can notice the involvement of attention mechanism in World Model does advance the performance. Based on the observation of curves, VMAV-C converges to stable results of rewards in high level more quickly, and workload of sampling from actual environments to achieve comparable stable optimal cumulative rewards is even less, suggesting the varying attention on historical information for prediction is reasonable in practice. The loss tendency in value network also tells VMAV-C can help state value network reach the optimal more easily than other two algorithms.

## 5.4 Policy Learning in Complete Virtual Environment

In this section, we would make use of MDN-RNN model trained after 50 epochs, which means cycling the process of **Algorithm2** in 50 times, and sample initial state to directly train agent in the virtual environment. Details on model configurations has been described in **Section 5.2**. Each iteration corresponds to a mini-sequence of up to 32 time step transitions' learning in the virtual environment. Meanwhile, some testing on learned policies would be performed in actual environment every 1000 iterations in training. To further investigate the characteristics in virtual environment, we operate parameter sensitivity analysis on VMAV-C model, including the length of mini-sequences used in RNN and randomness parameter τ of GMM's influence towards VMAV-C's performance. Notice, there is no process of sequentially interacting with the actual environments as **Step 4**, and the actual environment no longer provides information for policy improvement but functions as a testing bed for goodness of policy learned from the virtual environment every 1000 iterations.

### Influence of Length in Mini-sequences

Here, four scenarios corresponding to values of randomness parameter τ in GMMs in the list {0.6,0.8,1.0,1.2} are investigated in experiments. With randomness parameter fixed as some level, we compare the performance of two structures of VMAV-Cs. One is to learn dataset of mini-sequence in 32 time steps, and another is to learn that in 16 time steps. Thus, the structure of AVF varies with the length of mini-sequence. Results in **Fig9** show that though the virtual environment learned from 16 time steps mini-sequences enables agent to receive ideal cumulative rewards, it is still worse than that learned from mini-sequences of 32 time steps, and such gap increases with iteration number in four levels of τ. With respect to the prediction accuracy for ending state $d_t$, VMAV-C learned from mini-sequences of 32 time steps achieves average accuracy over 0.2% higher than that with 16 recurrent units after convergence in four scenarios.

Else, the distribution of 1500 episodes, which were sampled for the virtual environment training, are analyzed here, and we find episodes, whose length is less than 16 time steps, occupy a proportion of 38% and only 16% episodes are with length of over 32 time steps. As it turns out, longer mini-sequences for MDN-RNN training can capture more dynamics in transitions, so VMAV-C trained from mini-sequences of 32 time steps is able to approach actual environment more precisely, and this is not restricted to the volume of learning dataset.

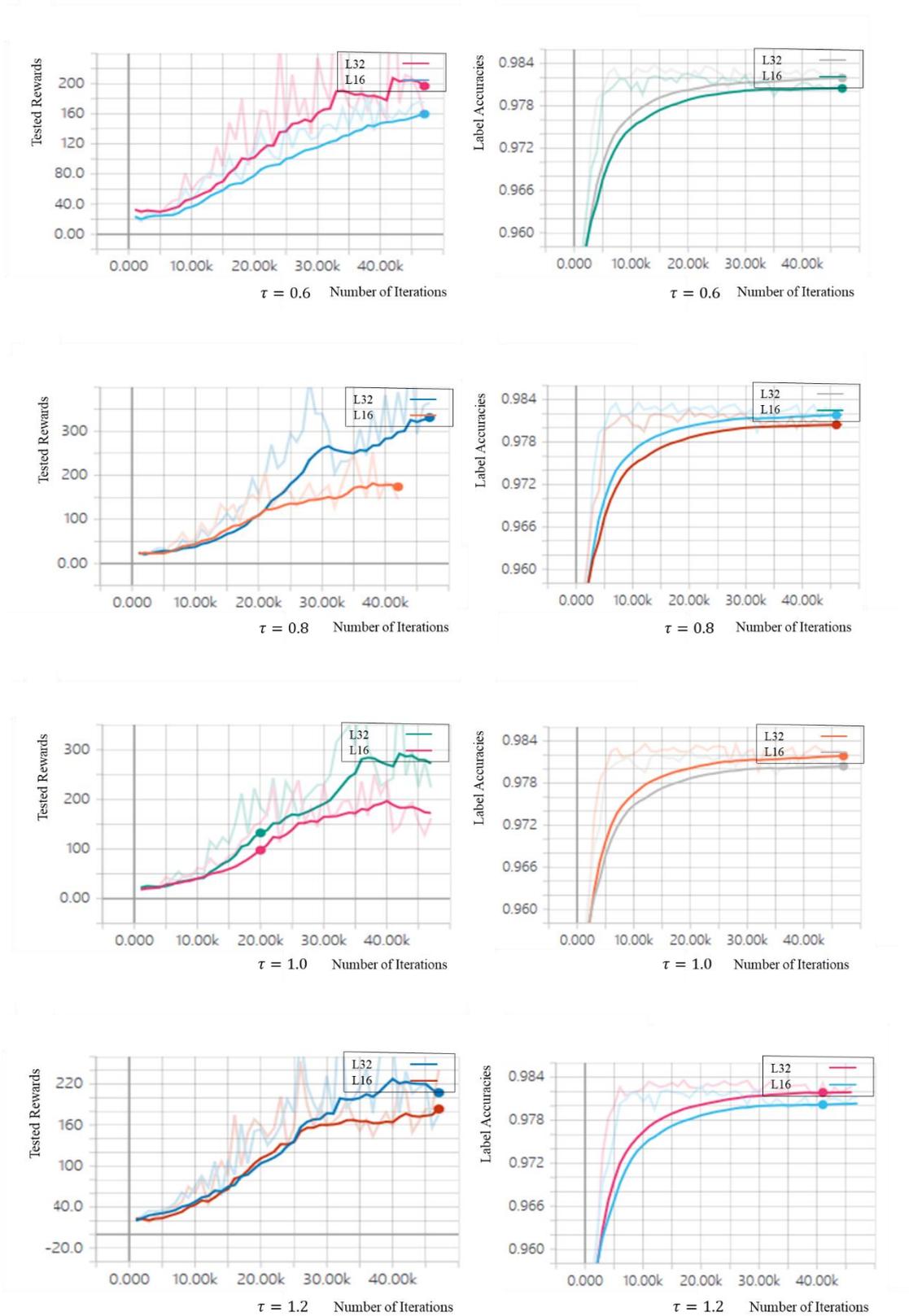

**Fig9. Performance Comparison of Two Referred Lengths of Mini-sequences.** Figures in the left are non-discount cumulative rewards, while the right ones are accuracies of predicting ending state in epochs. Smooth rates in Tensorboard are respectively 0.8 and 0.95 in tested rewards and label accuracies.

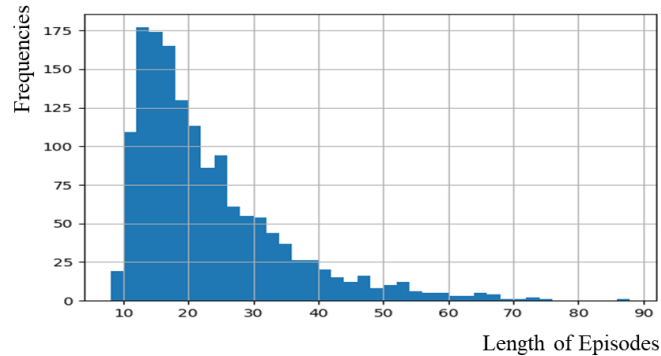

**Fig10. Distribution of Number of Time Steps in Initially Collected Episodes.**

## Influence of Randomness Parameter in GMMs

The investigation on the length of mini-sequences indicates the length is crucial in ending state of epoch prediction, and in this subsection the influence of randomness parameter τ in GMM towards VMAV-C performance is also studied. We map the collected results in **Fig9** to another view, focusing on the randomness parameter τ in GMMs. Here, the smooth rate in Tensorboard is chosen as 0.95 for illustration in **Fig11**. For the case of VMAV-C learned from mini-sequence in 16 time steps, we find higher τ values results in better cumulative rewards. For the case of VMAV-C learned from mini-sequence in 32 time steps, the optimal τ value for optimal cumulative rewards is in the range between 0.8 and 1. Similar to former analysis, it can be inferred that higher τ value aggravates the uncertainty of environment. Hence, for the VMAV-C learned from mini-sequence in 32 time steps, which can well learn the environment model, it would encounter overfitting with higher τ values. However, smaller τ value would leverage the robustness of model in some sense. So, the conclusion is the value of randomness parameter in GMMs should vary with power of learned environment model.

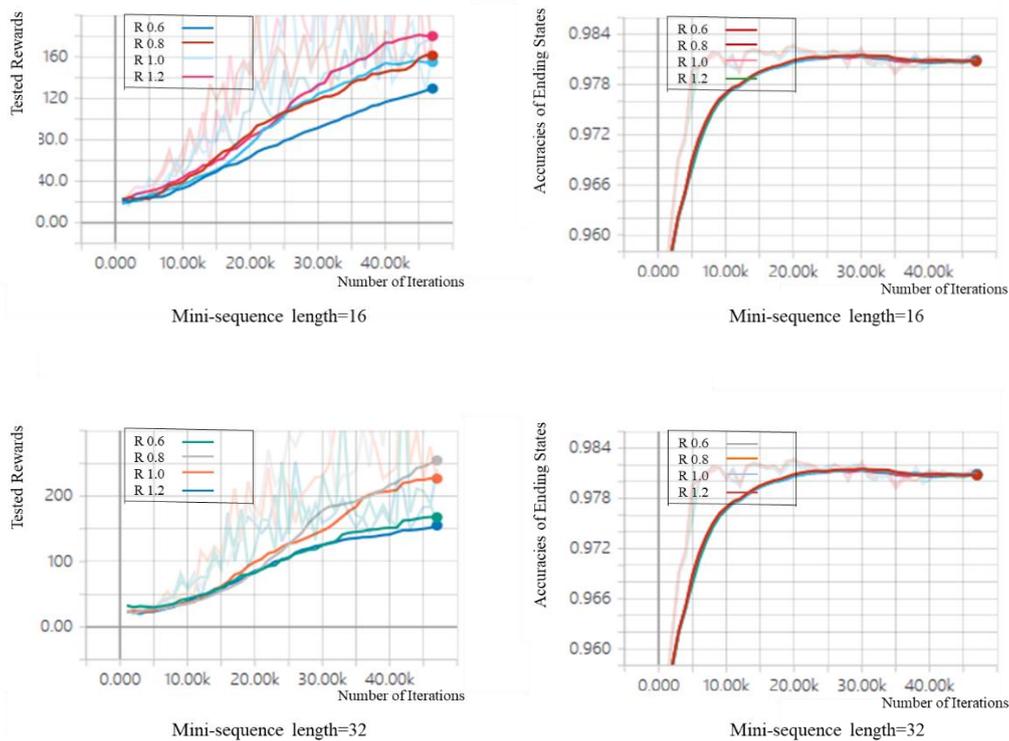

**Fig11. Performance Comparison in Various τ Values of GMMs.** The smooth rates in Tensorboard are respectively set as 0.9 and 0.95 for tested rewards and accuracies of ending states.

# 6. Conclussion

In this paper, we make modifications on former works about World Models and focused on improving policy learning procedure, including incorporating attention mechanism in estimates of state value and optimizing policy learning with PPO based AC algorithm. The experiment results demonstrate the effectiveness of these improvements, and further prove that limited experiences can still build task-beneficial virtual environment model with combination of VAE and MDN-RNN. In real environment, VMAV-C's performance is superior to former works, and agent trained in the virtual environment is capable of learning effective policy as well. Sensitivity analysis suggests that with appropriate parameter selection of MDN-RNN, the learned environment model can approach real environment model more precisely. In the future, we would explore methods to constructing more environment model and pay more attention to multi-agent system to boost simulation performance and efficiency.

## Acknowledgement

The authors declare no conflict of interest in this work and thanks to the sponsorship from National Science Foundation No.71701205 and No.71701206. Meanwhile, Dr. Qi Wang gratefully acknowledges the financial support from China Scholarship Council and supervision from Prof. Peter M. A. Sloot during his study in the Netherlands.